# Bridging the Gap between Local Semantic Concepts and Bag of Visual Words for Natural Scene Image Retrieval

Yousef Alqasrawi
*Faculty of Information Technology, Applied Science Private University, Amman, Jordan*

**Abstract:** This paper addresses the problem of semantic-based image retrieval of natural scenes. A typical content-based image retrieval system deals with the query image and images in the dataset as a collection of low-level features and retrieves a ranked list of images based on the similarities between features of the query image and features of images in the image dataset. However, top ranked images in the retrieved list, which have high similarities to the query image, may be different from the query image in terms of the semantic interpretation of the user which is known as the semantic gap. In order to reduce the semantic gap, this paper investigates how natural scene retrieval can be performed using the bag of visual word model and the distribution of local semantic concepts. The paper studies the efficiency of using different approaches for representing the semantic information, depicted in natural scene images, for image retrieval. An extensive experimental work has been conducted to study the efficiency of using semantic information as well as the bag of visual words model for natural and urban scene image retrieval.

## 1. INTRODUCTION

The rapid development of new information technologies, use of digital cameras in our daily lives, photo sharing websites and social networks, have led to an explosion in the amount of visual media such as images and videos available. Moreover, huge amount of images are collected every day in various fields such as digital books, digital art, medical imaging, aerial and satellite [1]. Unlike digital computers, humans are able to interpret the semantic content of images, which is still beyond the capabilities of computer vision systems. For a computer system, an image is not more than a matrix of pixel values which are summarized by low-level features such as color or texture features. For humans it is an image which contains what he/she can see, such as sky, water or forest. This makes visual content analysis of images a challenging problem in computer vision and other related fields, such as artificial intelligence and image data management [2].

One of the main and challenging problems in image content representation is the semantic gap [3]. It is the gap between low-level image features of the image content and the human perception. For example, in content-based image retrieval (CBIR) systems, the user provides a query image and he/she wants the system to retrieve images from the database that are visually similar to the query image. However, retrieved images that are considered by the system visually relevant images to the query image may not semantically relevant (the semantic gap). Much of the early work in image representation was to develop algorithms to extract different low-level features from image visual content, such as color, texture and shape. These features are used by the system to build the index of images in the database. However, these low-level features may be unrelated to the concepts expressed in the image, such as sky and grass (semantic information).

Extracting semantic information from image content is a field that tries to mimic the way human perception works, which is still a challenging and difficult task to accomplish in a computational system. One goal of image content analysis is to narrow the semantic gap between the image's visual content and the human understanding of image content.

Image retrieval is a challenging task in computer vision that is affected by this research problem when matching the semantics of images in the database. The use of image modelling based on local features has provided significant progress in terms of robustness, efficiency and quality of results. Much of the previous work on image representation was based on bag of visual words (BOW) model [2, 4, 5, 6, 7, 8, 9], a model that was brought from the text document retrieval field [10].

The bag of visual words model is based on local features extracted from image content using features detector and descriptor, such as difference of Gaussian (DoG) detector and scale-invariant feature transform (SIFT) descriptor [11]. These descriptors are then quantized into a number of clusters, called visual words, where an image is then represented as a histogram of the occurrence of each visual word in an image. This model has shown an impressive performance in image classification and object recognition problems. It is considered as an intermediate semantic representation of the image content.

This paper investigates how natural scene retrieval can be performed using the bag of visual word model and the distribution of local semantic concepts. An image is considered as a collection of local semantic concepts. These semantic concepts are semantic labels, such as sky and grass, used to describe the visual content of image regions. Representing the image content as a collection of local semantic concepts would make the image representation more semantically meaningful. These local semantic concepts appear in an image can be summarized as a histogram of their occurrence in the image. This histogram is called the concept-occurrence vector (COV).

Also, in this paper, the bag of visual words model can be considered as an intermediate semantic representation of the image content. Although no semantic labels are contained in the bag of visual word histograms, the visual words generated from the clustering process can be regarded as visual words of semantic information. This can be justified in the sense that similar local keypoints may be allocated to the same cluster (visual word), particularly in the case of using integrated visual vocabularies. This paper presents an extensive comparative study between local semantic concepts, bag of visual words, and other baseline methods, such as color and texture which represent the visual content of images without any semantic information for the task of natural scene image retrieval.

## 2. RELATED WORK

Content-based image retrieval has been introduced as a possible solution to overcome text-based image retrieval. In CBIR system, images are represented by their visual content, such as color and texture and a query image is introduced to the system which returns all similar images. This concept of query is called query-by-example (QBE). But due to the semantic gap that exist between low-level image features and high-level semantic understanding of images, CBIRs systems based on low-level features often fail to fulfill user demands [3, 12, 13, 14, 15]. In traditional CBIR, several systems have been developed for different domains and all of them have a common pipeline: extract descriptors from images and find similarities between a query image and images in the image collection. In general, Euclidean distance is used as a similarity measure between image descriptors. A comparison study between different low-level features for content-based image retrieval has been illustrated and analyzed for small datasets in [16] and for web scale images in [17]. The former study showed that color histogram can be used as a baseline for different CBIR applications. Nevertheless, they emphasized on the importance of semantic image analysis and understanding that has witnessed much work using semantic concepts [16]. They suggested that, for better content-based image retrieval, image descriptors needs to be combined with semantic concepts or textual information. This paper presents different approaches for integrating semantic concepts with low level features.

Many CBIR systems have used nearest neighbor approach to retrieve similar images. In QBIC system [18], color histograms, moment-based shape features and texture features are used to represent image contents. Another popular image retrieval system is Blobworld [19] developed at the UC Berkeley. In Blobworld, images are segmented into regions (blobs) using Expectation-Minimization (EM) algorithm clustering image pixel properties, color, texture and position information. An image is represented by color and texture features extracted from blobs rather than the entire image and the user selects a region and the system returns images composing similar regions. In SIMPLIcity [20] image retrieval system, images are segmented into regions using wavelet-features and k-means clustering algorithm. They have developed region-matching similarity metric to match all segmented regions automatically. They claimed that pre-classification of images enhances the retrieval results. Also, Iqbal and Aggrawal [21] developed another CBIR system, called CIRES, to improve image search results using image structure in combination with color histogram and Gabor features as texture.

In order to derive image content features which are semantically relevant to the user's perception, researchers in computer vision have focused on developing schemes which link image regions with semantic concepts. Liu et al. [12] categorized techniques to infer high-level semantic information into five types: (1) derive semantic concepts using Ontology, (2) derive semantic concepts using machine learning, (3) derive semantic concepts using relevance feedback, (4) derive semantic concepts using semantic templates, and (5) derive semantic concepts from textual information located with images for Web retrieval. . Most approaches are related to type (2) which adopt machine learning approaches to learn high-level semantic of images using visual content and textual features. For semantic-based image retrieval using machine learning algorithms images are first segmented into regions by fixed grid size, image segmentation or by using salient points. The next step is to associate semantic labels with image regions or with the entire image. This step is similar to image regions annotation and also used in image classification.

Sun and Ozawa [22] proposed a region-based image retrieval approach using wavelet transform. Image regions are obtained by clustering the wavelet coefficients in the Low-Low frequency sub-band of image wavelet transform. Another approach for region-based image retrieval was proposed by Chen and Wang [23]. Images are first segmented into regions, each of which is characterized by a fuzzy feature representing color, texture and shape features. An image is then represented as a set of fuzzy sets corresponding to image regions. To improve the performance of image retrieval, Chen et al. [24] proposed a cluster-based image retrieval which includes the similarity information between retrieved images. Some of the recent works on semantic-based image retrieval based on bag of visual words has been presented Nister and Stewenius [25] and Chen et al. [26]. Vieux et al. [27] proposed a similar idea to the BOW model, called bag of regions model. They proposed incremental clustering algorithm for building visual vocabularies. They have shown promising results on some of the public datasets. In contrast to [26] and [27], the work in this paper focuses on reducing the semantic gap by using bag of visual words with different configurations and the distribution of local semantic concepts to retrieve images that are similar to the query image.

## 3. IMAGE RETRIEVAL BASED ON ANNOTATED IMAGE REGIONS

In this section, a query image can be considered as a collection of local semantic concepts, each of which describes a particular region in the image. Thus, image regions in the database are assumed to be annotated with semantic concepts. For this reason, the natural scene dataset presented in Section 5 (Dataset1) which provides ground truth data about annotations of image regions is employed. Images in the dataset are divided into 10×10 regular grid which yields 100 regions per image. All these regions were manually annotated with semantic concepts. These semantic concepts are *sky*, *water*, *grass*, *trunks*, *foliage*, *field*, *rocks*, *flowers,* and *sand*. The bag of visual words is used to represent the visual content of these regions [32].

We follow our previous work [32, 33] to represent image regions where DoG detector and SIFT descriptor are employed to find and represent local keypoints in all images of the dataset. Each SIFT descriptor is a vector of size 128-D. For each natural scene category, k-means algorithm is applied to all descriptors to build visual vocabularies. These vocabularies are then aggregated to form an integrated visual vocabulary of size (K×M) where K is the vocabulary size and M is the number of scene categories. Bag of visual words is used to represent image regions. A frequency histogram is generated from each image region where the number at each bin corresponds to the frequency of occurrence of each visual word in that image region. To annotate regions of test images, SVM and KNN classifiers are trained on different configurations of the bag of visual words model.

To improve the power of visual vocabulary, we have incorporated spatial information about local keypoints when clustering their features. Natural scene images, such as coasts, contain semantic concepts that usually appear in common places. For example, the sand concept can be found at the bottom of an image whereas the sky and water concepts appear at the top. Also, building visual vocabularies using k-means algorithm may lead to group dissimilar local keypoints in the same cluster. To avoid this problem, images are partitioned into two halves of equal size: upper and lower. An image of size W×H is divided into two halves, each has a dimension of (W/2×H/2). In this case, two integrated visual vocabularies are generated: Upper integrated visual vocabulary and Lower integrated visual vocabulary. Upper integrated visual vocabulary is generated by clustering SIFT features located at the upper half of all training images. Lower integrated visual vocabulary is generated by clustering SIFT features located at the lower half of all training images. The preliminary experimental work demonstrates that the SVM classifier achieves the best results to annotate image regions at image halves, i.e. upper halve and lower halve.

To this end and having that test images are annotated with local semantic concepts, an image is described as a collection of local semantic concepts. These semantic concepts provide semantic interpretation of the image content which can reduce the semantic gap between the user perception and the image content. However, different images would have different number of local semantic concepts. Thus, it is important to summarize the amount of local semantic concepts found in an image into a global feature vector. To do so, the concept-occurrence vector (COV) proposed by Vogel and Schiele [34] is adopted. To represent an image, the frequency of occurrence of each semantic concept is determined. Since there are nine semantic concepts available in the natural scene dataset, each image can be represented as a feature vector of size nine. Each component corresponds to a semantic concept and it contains an integer number of the frequency of occurrence of this semantic concept in the image. Figure 1 shows image representation using the COV adopted from [34]. This figure shows an image divided into 10×10 regular grid and each region is manually annotated with one of the nine semantic concepts. The COV is then generated by counting how many times a particular semantic concept appears in the image. For example, the semantic concept sky appears 47 times and a half. By dividing this number by 100 yields a normalized occurrence value of 47.5%. This is replicated for all semantic concepts. All natural scenes in the database are indexed using their COV generated from the ground truth annotations.

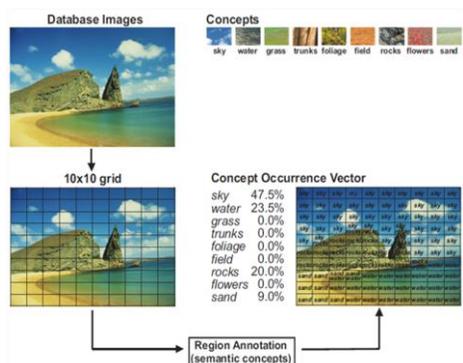

Figure 1: Image representation using concept-occurrence vector (COV) [34]

To retrieve images from the database, the COV is first constructed from the query image. This feature vector is then compared to all COVs of images available in the database using a similarity measure, such as the Euclidean distance, and a ranked list of the relevant images are retrieved as a response to the query image. The performance of using COVs generated from ground truth annotations for natural scene retrieval will be presented in the experimental work section. Their retrieval results will be considered as benchmark which gives the best retrieval results to expect.

Different configurations of the bag of visual words model were employed to represent the visual characteristics of image regions. The task was to annotate test images with local semantic concepts assigned to image regions. We have conducted 14 experiments using the SVM classifier and 14 different image region representation approaches.

Figure 2 shows the main steps needed to construct the COV for a new/test image. As mentioned above, there are 14 different approaches to represent the visual content of image regions (see * note in Figure 2 for the list of approaches used in this paper). The COVs resulted from each approach will be compared in the experimental work section to see which of these approaches are suitable to represent image regions for the task of natural scene image retrieval. Their retrieval performances will be compared to the COVs benchmark.

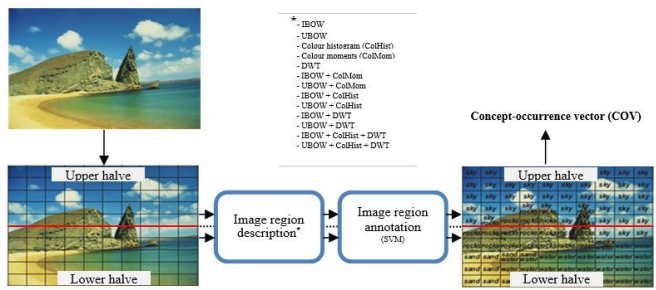

Figure 2: Image representation using COV of local semantic concepts assigned to image regions at the upper and lower halves of the image.

## 4. IMAGE RETRIEVAL USING BAG OF VISUAL WORDS

The problem of using global descriptors, such as color histograms, to represent the visual content of an image is that they cannot describe different characteristics of the image content. To overcome this problem, descriptors of local keypoints aim to capture characteristics of different parts in an image. Grouping these descriptors into visual words, using a clustering algorithm, may result in allocating these descriptors to visually similar visual words.

The bag of visual word model has shown to be effective to represent the distribution of local keypoints, detected in the image, for the natural scene classification task [33]. Descriptors of local keypoints mainly represent the intensity information extracted from regions around interest points while discarded color information. Also, bag of visual word model ignores the order of local keypoints found in the image. Moreover, clusters (visual words) of the traditional visual vocabulary do not capture the differences between local keypoints found in different scene categories. Thus, in [33], different approaches were introduced to overcome these limitations and have been applied to the natural scene classification task. Some of these approaches will be used in this paper to investigate how well they perform for the natural scene retrieval task. Also, this paper aims to compare the performance of using the COV and the BOW for representing the semantic information of images to perform the retrieval task. The difference between using the COV and the bag of visual words model for natural scene retrieval is that the former approach requires images to be manually annotated with local semantic concepts in order to train semantic concept annotators which are used to annotate new image, while the later approach does not. However, the dimensionality if COV is too small compared to the high dimensional BOW histograms.

In this section, no annotations are required to represent the semantic information in the image. Images in the database are assumed to be indexed by their BOW histograms and with different configurations.

Table 1: Image representation using 14 different approaches. The second column describes each approach and specifies the size needed for each representation.

| Image representation approach | Description of the approach and the dimensionality of feature vectors produced |
|---|---|
| ColHist | HSV colour histogram (H: 36, S: 32, V: 16 = 84-D), for grey images (36-D) |
| PColMom_L0 | HSV pyramidal colour moment at level 0 (firs & second moments = 6-D) |
| DWT | Discrete Wavelet transform (H: 6-D, S: 6-D, V:6-D = 18-D) |
| ColHist+DWT | Colour histograms integrated with DWT (102-D) |
| PColMom_L2 | HSV pyramidal colour moments at level 2 (126-D) |
| UBOW | BOW histogram using universal visual vocabulary (200-D) |
| IBOW | BOW histogram using integrated visual vocabulary (# scene categories ×200) |
| PUBOW_L1 | Pyramidal UBOW at level 1, (200×5 = 1000-D) |
| PUBOW_L2 | Pyramidal UBOW at level 2, (200×21 = 4200-D) |
| PUBOW_L2+PColMom_L2 | Pyramidal UBOW at level 2 + Pyramidal colour moments at level 2 ((200-D+6-D)×21 = 4326-D) |
| PIBOW_L1 | Pyramidal IBOW at level 1 (# scene categories × 200-D × 5 ) |
| PIBOW_L2 | Pyramidal IBOW at level 2 (# scene categories × 200-D × 21) |
| PIBOW_L2+PColMom_L2 | Pyramidal IBOW at level 2 + Pyramidal colour moments at level 2 ((# scene categories × 200-D × 21 + (6-D×21)) |
| PIBOW_L2+WPColMom_L2 | Pyramidal IBOW at level 2 + Weighted Pyramidal colour moments at level 2 ((# scene categories × 200-D × 21 + (6-D×21)) |

Table 1 shows details of the 14 different approaches used in this paper to represent the visual content of images. Most of these image representations are obtained from the work presented in [33] for three natural scene datasets with 6, 8, and 15 scene categories, respectively. The next section introduces the experimental work and results of natural scene retrieval using all approaches presented in this section and the previous one.

## 5. EXPERIMENTAL WORK

### 5.1. Evaluation Methodology

There are different measures for retrieval performance proposed in the information retrieval and pattern recognition literature. They are used to evaluate the results of the experiments. The most common and widely used performance measures in information retrieval are the precision P and recall R. These measures evaluate how well an information retrieval system performs on the ground truth data. It gives a good indication of the image retrieval system performance [28].

For a given query Q, let X be the number of relevant images that belongs to the same category in the image dataset and Y be the set of all images retrieved. Assume that Z be the number of retrieved images coming from the same category (i.e., correctly retrieved images) which are among the Y retrieved images, then precision P and recall R care defined as:

$$P = \frac{Z}{Y} \qquad (1)$$

$$R = \frac{Z}{X} \qquad (2)$$

These two values are commonly combined into a so called recall-precision graph where precision values (y-axis) are plotted against recall values (x-axis) and each dot in the graph represents a retrieved image of the ordered result list. It shows how many retrieved images retrieved are relevant or irrelevant among the top ranked images. However, interpreting recall-precision graphs is not an easy task. Thus, it is possible to summarize precision and recall in a single value by calculating the average precision at each point when a relevant image is found and then calculate the mathematical mean of these precisions. This measure is called the Mean Average Precision (MAP). For every query image, precision and recall measures are computed over all images retrieved which are ranked, in a descending order, according to a similarity measure. Then the measures are averaged over all the queries in the test dataset.

### 5.2. Image Datasets

There are many image datasets available in the computer vision literature, but most of them are dedicated to object detection and categorization tasks. Performance of the proposed approaches for image retrieval are tested on two types of image datasets: a dataset with natural scene images only, which is our main concern, and datasets with heterogeneous images including different kind of images, such as urban images. The reason for choosing natural scene images is that they generally are difficult to categorize in contrast to object-level classification because natural scenes constitute a very heterogeneous and complex stimulus class [29]. Also, we considered scene images that constitute artificial objects to allow fair and straightforward comparison with state-of-the-art scene classification methods. Four datasets were used in our experiments (see Figure 3):

*Dataset 1*: This dataset, kindly provided by Vogel et al. [29], contains natural scene images only with no man-made objects. It contains a total of 700 color images of resolution 720×280 and distributed over 6 categories. The categories and number of images used are: coasts, 142; rivers/lakes, 111; forests, 103; plains, 131; mountains, 179; sky/clouds, 34. One challenge in this image dataset is the ambiguity and diversity of inter-class similarities and intra-class differences which makes the retrieval task more challenging.

*Dataset 2*: This dataset is a subset of the Oliva and Torralba [30] dataset. It constitutes images of natural scene categories with no artificial objects, which are semantically similar to images in Dataset1 and is distributed as follows: coasts, 360; forest, 328; mountain, 374; open country, 410. The total number of images in this dataset is 1472.

*Dataset 3*: This dataset contains heterogeneous image categories. It consists of a total of 2688 color images, 256x256 pixels, and distributed over 8 outdoor categories. The categories and number of images used are: coast, 360; forest, 328; mountain, 374; open country, 410; highway, 260; inside city, 308; tall building, 356; street, 292. This dataset is created by Oliva and Torralba [30] and is available online at http://cvcl.mit.edu/database.htm.

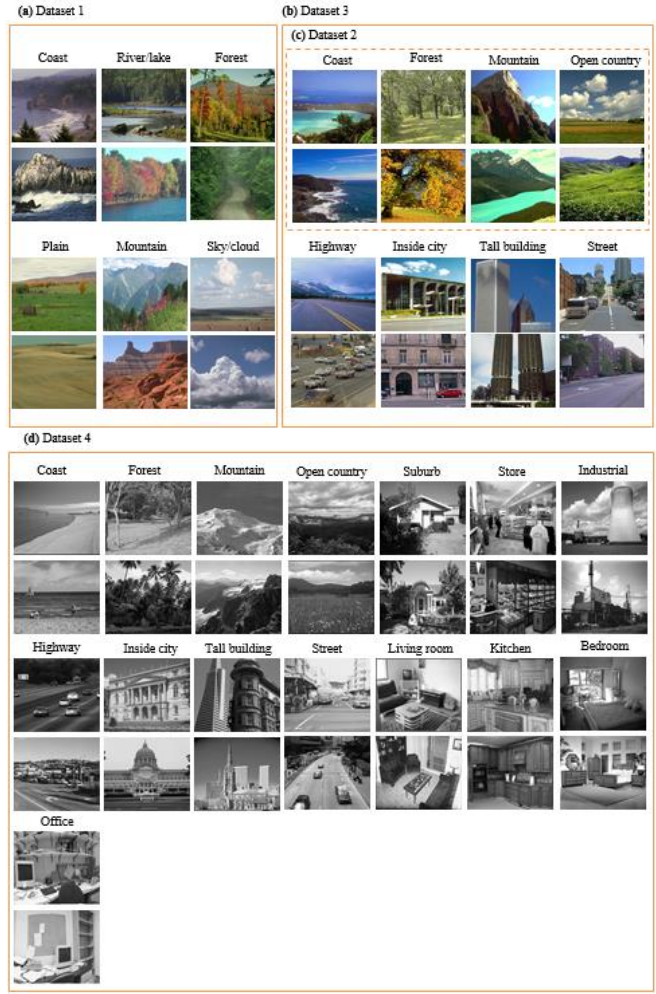

Figure 3: Some examples of the images used for each category from the Dataset 1, Dataset 2, Dataset3 and Dataset 4 respectively.

*Dataset 4*: This dataset, provided by Lazebnik et al. [31], contains 15 heterogeneous natural scene image categories. All images in this dataset are grayscale images (i.e., no color images). It contains different kind of images and the average image size is 300x250 pixels. Images are distributed over categories as follow: highway, 260; inside city, 308; tall building, 356; street, 292; suburb, 241; forest, 328; coast, 360; mountain, 374; open country, 410; bedroom, 216; kitchen, 210; living room, 289; office, 215; industrial, 311; store, 315.

**5.3. Experimental setup**

To be able to compare the performance of different image retrieval algorithms, ground truth images are used, i.e. images in the dataset should be grouped into categories, thus every image in the dataset belongs to one of the predefined scene categories. All experimental works presented in this paper are evaluated based on 10-folds cross-validation. For a particular image dataset, 10% of the images are randomly selected from each scene category. These images are used as queries in the image retrieval experiments. The other remaining 90% of the images from each scene category form the ground truth images from which images are retrieved in response to the query images.

Different measures are used to evaluate the performance of the different image retrieval implementations. Firstly, the recall-precision graphs are used for each approach. For each scene category, ranked precisions and recalls of all query images (test images) are averaged and plotted on the recall-precision graph. Another performance measure is the mean average precision (MAP). This measure is calculated for each scene category. The third measure is the retrieval accuracy of each approach which is the arithmetic mean of the MAPs over all scene categories. For all image representation approaches, each image is represented as a vector of values which is normalized to a unit length. For multiple feature image representation, to be concatenated, each feature type is first normalized. The Euclidean distance is used to find similarities between the query image and images in the database.

Three natural scene datasets are used in this section to evaluate the retrieval performances of different image representation approaches. The first dataset, referred to as Vogel_6DS, consists of 700 color natural scene images distributed over six scene categories [29]. The second dataset, referred to as Oliva_8DS, consists of 2688 color images distributed over 8 scene categories [30]. The third dataset, referred to as Lazebnik_15DS, contains 4485 gray images distributed over 15 scene categories [31].

### 5.4. Experiments on image retrieval using COV

This section presents the experimental work for image retrieval using the concept-occurrence vector evaluated on the dataset Vogel_6DS. In this section a set of experiments are carried out. The first experiment is carried out to evaluate the performance of using COV, constructed from the ground truth annotations of image regions, for the image retrieval. The results obtained from this experiment serves as a benchmark to evaluate how discriminative are the bag of visual word model and other baseline methods in describing image regions, which in turn used by concept annotator to generate local semantic concept needed to construct the COV.

The MAP results of each scene category using the COV benchmark is depicted in the first row of Table 3. The MAPs for scene categories Coasts and River/lakes are the most difficult categories to retrieve. Images from both scene categories are visually ambiguous. The retrieval accuracy of using the COV benchmark is 83%. The same experiment was carried out in [29] and they achieved 80.6% retrieval accuracy. The difference between their experiment and our experiment is that they used the SVM classifier to rank the retrieved images while in this work the retrieved images are ranked using the Euclidean distance. Thus, to make fair comparisons, the results obtained in this section are used as a benchmark to compare the retrieval performances of using the COV obtained using other approaches. The recall-precision graph of the COV benchmark is also depicted in Figure 4 (a).

In the next experiments, the COVs are generated from image regions labelled with local semantic concepts using the different image representation approaches that gave the best annotation results in our preliminary experiments which has been mentioned in this paper. These approaches are listed in Table 2. Each approach is used to annotate images with local semantic concepts. From these local semantic concepts the COVs are generated as global image representation. The COVs are then used in the retrieval task. It is worth to remind that the universal bag of visual words (UBOW) and integrated bag of visual words (IBOW) listed in Table 2 are concept-based bag of visual words (CBOW) generated at the upper and lower halves of images using the universal and integrated visual vocabularies.

For simplicity reasons, the UBOW and IBOW are used in this paper to refer to the CBOW generated either by the universal visual vocabulary or the integrated visual vocabulary.

Table 2: Different approaches used to represent image regions. These approaches are used to annotate image regions with local semantic concepts.

| | | | |
|---|---|---|---|
| (1) | Colour histogram (ColHist) | (2) | Colour moments (ColMom) |
| (3) | DWT | (4) | Colour histogram + DWT |
| (5) | UBOW | (6) | IBOW |
| (7) | UBOW + Colour histogram | (8) | UBOW + Colour moments |
| (9) | UBOW + DWT | (10) | UBOW + Colour histogram + DWT |
| (11) | IBOW + Colour histogram | (12) | IBOW + Colour moments |
| (13) | IBOW + DWT | (14) | IBOW + Colour histogram + DWT |

The MAPs results of each scene category using the 14 approaches listed in Table 2 are shown Table 3. The results can be compared directly to the retrieval results obtained using the COV benchmark. The color histogram shows a good retrieval performance (72%) compared with the color moments (58%) and DWT (65%) with a slight improvement when concatenated with the DWT.

Table 3: The MAPs of each scene category using the COV benchmark and the other 14 different approaches. The last column shows the retrieval accuracy of each of the corresponding approach.

| | MAP per scene category | | | | | | |
|---|---|---|---|---|---|---|---|
| | Coasts | River/lakes | Forests | Plains | Mountains | Sky/clouds | Acc. |
| COV | **0.75** | **0.63** | **0.95** | **0.77** | **0.86** | **0.99** | **0.83** |
| ColHist | 0.65 | 0.46 | 0.90 | 0.60 | 0.76 | 0.91 | 0.72 |
| ColMom | 0.56 | 0.33 | 0.81 | 0.45 | 0.62 | 0.69 | 0.58 |
| DWT | 0.69 | 0.38 | 0.81 | 0.42 | 0.66 | 0.96 | 0.65 |
| ColHist+DWT | 0.68 | 0.49 | 0.93 | 0.65 | 0.75 | 0.90 | 0.73 |
| UBOW | 0.64 | 0.39 | 0.81 | 0.53 | 0.74 | 0.87 | 0.66 |
| IBOW | 0.72 | 0.45 | 0.89 | 0.61 | 0.79 | 0.97 | 0.74 |
| UBOW+ColHist | 0.75 | 0.59 | 0.95 | 0.66 | 0.84 | 0.99 | 0.80 |
| UBOW+ColMom | 0.75 | 0.50 | 0.92 | 0.58 | 0.81 | 0.97 | 0.75 |
| UBOW+DWT | 0.73 | 0.46 | 0.93 | 0.59 | 0.82 | 0.99 | 0.75 |
| UBOW+ColHist+DWT | 0.73 | 0.57 | 0.93 | 0.67 | 0.85 | 1.00 | 0.79 |
| IBOW+ColHist | 0.75 | 0.60 | 0.95 | 0.67 | 0.84 | 0.99 | 0.80 |
| IBOW+ColMom | 0.77 | 0.56 | 0.95 | 0.58 | 0.82 | 0.97 | 0.78 |
| IBOW+DWT | 0.76 | 0.49 | 0.95 | 0.58 | 0.84 | 0.99 | 0.77 |
| IBOW+ColHist+DWT | 0.74 | 0.58 | 0.95 | 0.67 | 0.85 | 1.00 | 0.80 |

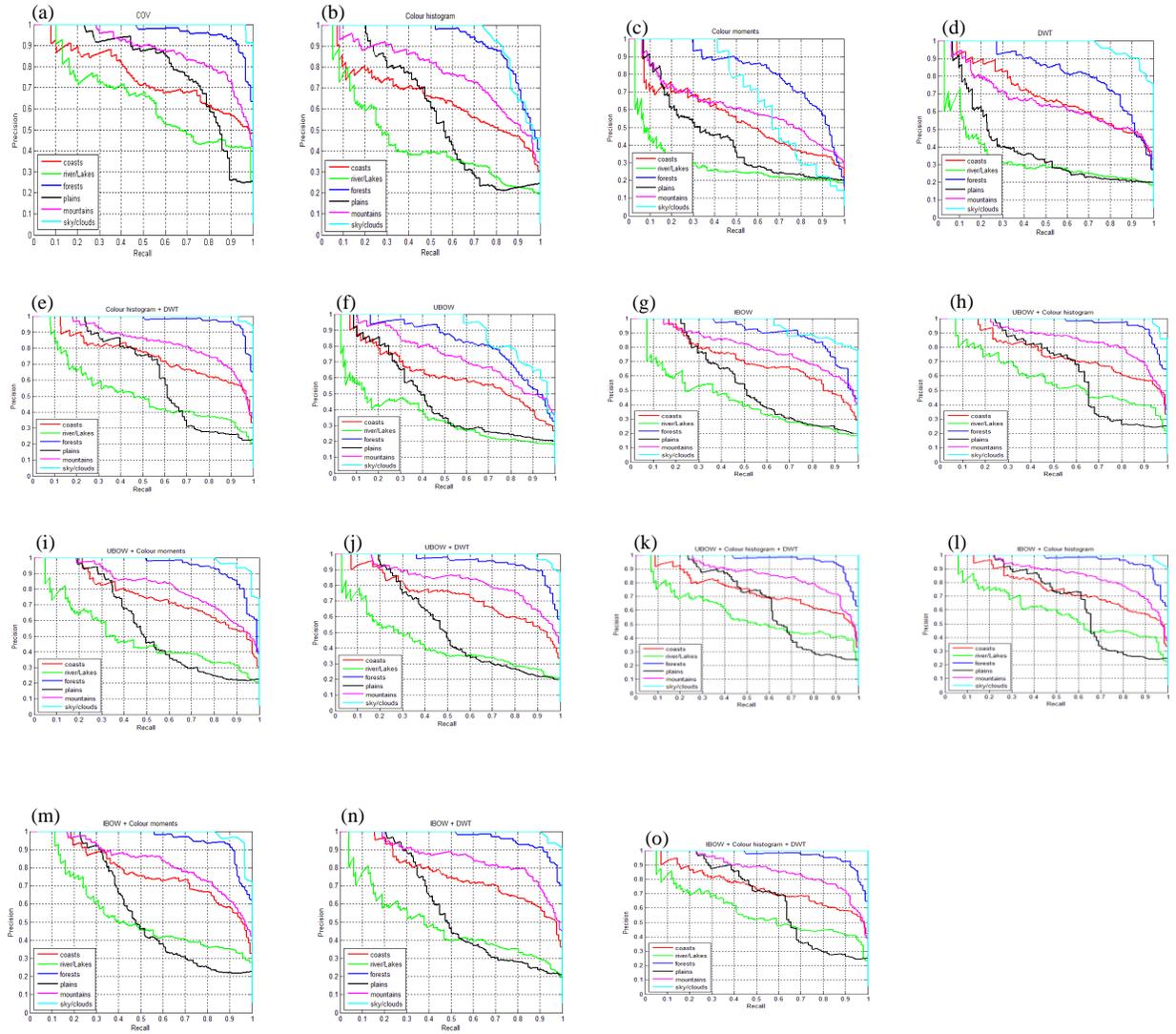

Figure 4: Precision-recall graphs, for Vogel_6DS, using COV image representation implemented using the ground truth annotations (a) and annotations obtained by different region representation approaches (b-o).

Interestingly, the color histogram achieved better retrieval accuracy than the UBOW. The reason for the worse retrieval performance of the UBOW approach is due to that the universal visual vocabulary used to build the UBOW is not discriminative enough. Also, color information is not included with UBOW.

The UBOW has gained better performance (80%) when it is concatenated with the color histogram, UBOW+ ColHist. The IBOW approach shows better performance than the UBOW and color histogram and also has improved when combined with the color histogram. The retrieval results of the 14 experiments revealed that the BOW model combined with the color information gained very good retrieval results (80%) compared to the retrieval results of the COV benchmark (83%). The recall-precision graphs of the 14 experiments are shown in Figure 4 (b-o). From this figure, it is possible to see the differences in the retrieval performance of each scene category and using the 14 different approaches.

The recall-precision plots of each scene category are averaged such that the performance of each approach can be visualized as a recall-precision graph of all approaches. This can be seen in Figure 5. It is obvious that the COVs based on BOW models can perform closer to the COVs benchmark.

### 5.5. Experiments on image retrieval using BOW

This section presents the experimental results of using the different approaches listed in Table 1 and evaluated on the three natural scene datasets Vogel_6DS, Oliva_8DS and Lazebnik_15DS for natural scene retrieval. The experimental work presented in this section assumes no annotations are available for the natural scenes. The only information available about images is their scene category.

This section tries to answer the following questions. What is the performance of using bag of visual words model for natural scene retrieval? How good is the spatial pyramid bag of visual words model for the natural scene image retrieval? What is the effect of using our proposed weighting approaches presented in [33] on the performance of scene retrieval? How good these approaches are compared to the baseline methods? These questions can be answered by evaluating the performances of all approaches presented Table 1 for natural scene retrieval. Next, three sets of experiments are presented in the following subsections each of which corresponds to experiments carried out using a particular natural scene dataset.

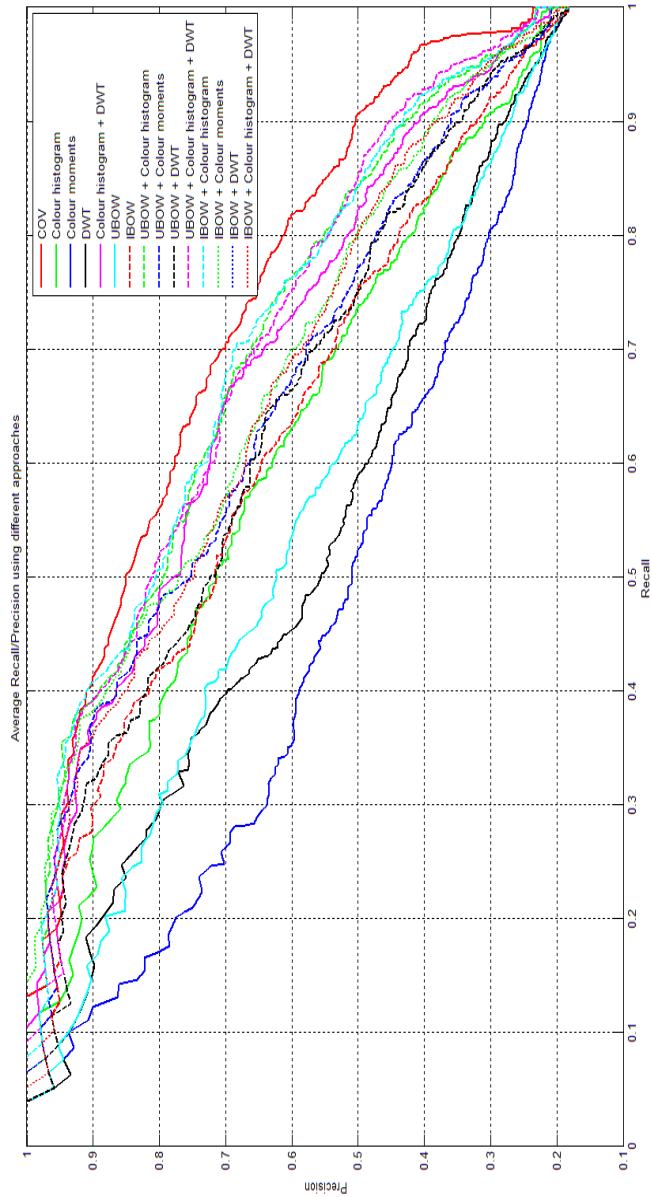

Figure 5: Recall-precision graph, for Vogel_6DS, of the performance of the COV benchmark and the 14 different approaches.

### 5.5.1 Experimental results: Vogel_6DS dataset

This section presents the experimental results of using different configurations of the bag of visual words model to represent the semantic information contained in natural scene images for the natural scene retrieval task. Other baseline methods are also used for comparisons such as color histogram and DWT. Color histogram and DWT are extracted from the entire image as a global representation for that image. The pyramidal color moments and all configurations of the bag of visual word models were illustrated in [33]. Thus, these approaches will not be explained again.

The performance of using bag of visual words to represent image content has gained better retrieval performance over the baseline methods. The IBOW shows better performance than the UBOW. Also, using the spatial pyramid layout has gained another improvement in the retrieval performance over using the UBOW and IBOW without any spatial information. Some scene classes have gained better retrieval performances when the weighting approach is used to add color to the pyramidal IBOW model.

The performances of all experiments presented in this section are shown in Table 4. The pyramidal integrated bag of visual words integrated with the weighed color moments both implemented at level 2 achieved the best retrieval performance with an increase of (+16%) over the color histogram and (+13%) over the UBOW. The color moments perform surprisingly well compared with the color histogram and DWT. The retrieval accuracy of scene categories Sky/clouds, Plains and Forests have gained batter performance compared to using only baseline methods. To compare the behavior of the different image representation approaches in the retrieval task, the recall-precision plots of all scene categories presented in each recall-precision graph are averaged into a single recall-precision plot.

Table 4: The MAPs of each scene category, for Vogel_6DS, using the COV benchmark and the other 14 different approaches presented in Table 1. The last column shows the retrieval accuracy of each of the corresponding approach.

|  | MAP per scene category |  |  |  |  |  |  |
|---|---|---|---|---|---|---|---|
|  | Coasts | River/lakes | Forests | Plains | Mountains | Sky/clouds | Acc. |
| **COV** | **0.75** | **0.63** | **0.95** | **0.77** | **0.86** | **0.99** | **0.83** |
| ColHist | 0.53 | 0.44 | 0.61 | 0.38 | 0.57 | 0.39 | 0.49 |
| PColMom_L0 | 0.58 | 0.41 | 0.63 | 0.46 | 0.56 | 0.54 | 0.53 |
| DWT | 0.79 | 0.37 | 0.59 | 0.34 | 0.48 | 0.39 | 0.44 |
| ColHist+DWT | 0.55 | 0.44 | 0.62 | 0.39 | 0.59 | 0.35 | 0.49 |
| PColMom_L2 | 0.55 | 0.40 | 0.73 | 0.51 | 0.52 | 0.55 | 0.54 |
| UBOW | 0.54 | 0.32 | 0.58 | 0.35 | 0.60 | 0.75 | 0.52 |
| IBOW | 0.57 | 0.34 | 0.72 | 0.40 | 0.62 | 0.78 | 0.57 |
| PUBOW_L1 | 0.52 | 0.34 | 0.65 | 0.36 | 0.60 | 0.76 | 0.54 |
| PUBOW_L2 | 0.51 | 0.34 | 0.69 | 0.37 | 0.61 | 0.75 | 0.54 |
| PUBOW_L2+PColMom_L2 | 0.59 | 0.36 | 0.76 | 0.42 | 0.63 | 0.79 | 0.59 |
| PIBOW_L1 | 0.56 | 0.33 | 0.77 | 0.43 | 0.64 | 0.78 | 0.58 |
| PIBOW_L2 | 0.55 | 0.32 | 0.78 | 0.44 | 0.63 | 0.78 | 0.58 |
| PIBOW_L2+PColMom_L2 | 0.62 | 0.36 | 0.82 | 0.46 | 0.63 | 0.81 | 0.62 |
| PIBOW_L2+WPColMom_L2 | 0.64 | 0.45 | 0.84 | 0.48 | 0.63 | 0.83 | 0.65 |

The recall-precision plots of the retrieval performance of all image representation approaches are shown in Figure 6. It is obvious that all approaches work worse than the COV benchmark. This is due to the fact that the COV approaches rely on the local semantic concepts which require image regions to be annotated by the user. If images in the database are not annotated at region level, then the bag of visual words model becomes a good choice for natural scene retrieval since it demonstrated better retrieval accuracy than the baseline methods.

### 5.5.2 Experimental results: Oliva_8DS

This section presents the experimental results carried out on the Oliva_8DS dataset. The same approaches used in the previous section are also employed in this section but for different dataset. Images in the dataset used in this section are not annotated at image regions. Thus, it is not possible to compare the performance of BOW-based approaches against the COV approach. As mentioned at the beginning of this paper, the BOW approach can be considered as an intermediate semantic representation of the visual content of images. Compared to baseline methods, the BOW-based approaches achieved better retrieval performances in all scene categories.

Using the proposed BOW-based approaches, most scene categories achieved better retrieval results than the baseline methods. It is worth to note that natural scenes with man-made objects (Urban images), such as Street and Inside city achieved very good performances compared to scene categories without man-made objects. This can be justified by the ability of the SIFT features to capture the structure if buildings and other man-made objects. For natural scene categories without man-made objects the task of image retrieval becomes harder. The recall-precision graph obtained by averaging the recall-precision plots of all scene categories for each approach is depicted in Figure 7.

Table 5: The MAPs of each scene category, for Oliva_8DS, using the 14 different approaches presented in Table 1. The last column shows the retrieval accuracy of each of the corresponding approach.

|  | MAP per category |  |  |  |  |  |  |  |  |
|---|---|---|---|---|---|---|---|---|---|
|  | Coast | Forest | Highway | Inside city | Mountain | Open country | Street | Tall building | Acc. |
| ColHist | 0.29 | 0.65 | 0.47 | 0.44 | 0.32 | 0.40 | 0.57 | 0.37 | 0.44 |
| PColMom_L0 | 0.32 | 0.52 | 0.56 | 0.33 | 0.37 | 0.45 | 0.46 | 0.34 | 0.42 |
| DWT | 0.50 | 0.75 | 0.61 | 0.46 | 0.33 | 0.45 | 0.53 | 0.49 | 0.52 |
| ColHist+DWT | 0.30 | 0.68 | 0.49 | 0.45 | 0.33 | 0.40 | 0.58 | 0.37 | 0.45 |
| PColMom_L2 | 0.34 | 0.60 | 0.68 | 0.47 | 0.36 | 0.56 | 0.54 | 0.45 | 0.50 |
| UBOW | 0.52 | 0.86 | 0.48 | 0.76 | 0.65 | 0.47 | 0.67 | 0.47 | 0.61 |
| IBOW | 0.64 | 0.89 | 0.51 | 0.81 | 0.64 | 0.46 | 0.70 | 0.56 | 0.65 |
| PUBOW_L1 | 0.53 | 0.90 | 0.46 | 0.77 | 0.64 | 0.51 | 0.69 | 0.48 | 0.62 |
| PUBOW_L2 | 0.53 | 0.91 | 0.45 | 0.78 | 0.63 | 0.50 | 0.69 | 0.49 | 0.62 |
| PUBOW_L2+PColMom_L2 | 0.52 | 0.90 | 0.63 | 0.74 | 0.62 | 0.62 | 0.69 | 0.56 | 0.66 |
| PIBOW_L1 | 0.65 | 0.91 | 0.50 | 0.82 | 0.62 | 0.48 | 0.72 | 0.56 | 0.65 |
| PIBOW_L2 | 0.65 | 0.91 | 0.50 | 0.82 | 0.61 | 0.48 | 0.71 | 0.56 | 0.65 |
| PIBOW_L2+PColMom_L2 | 0.52 | 0.87 | 0.69 | 0.75 | 0.55 | 0.62 | 0.69 | 0.58 | 0.66 |
| PIBOW_L2+WPColMom_L2 | 0.57 | 0.87 | 0.69 | 0.77 | 0.59 | 0.64 | 0.77 | 0.62 | 0.69 |

The precision plot of our proposed approach (approach number 14 in the figure) shows always better retrieval accuracy than other approaches. The MAP results of each scene category can be shown in Table 5. This table shows that the PIBOW_L2+WPColMom_L2 approach reports (69%) compared with traditional UBOW which has achieved (61%) retrieval rate.

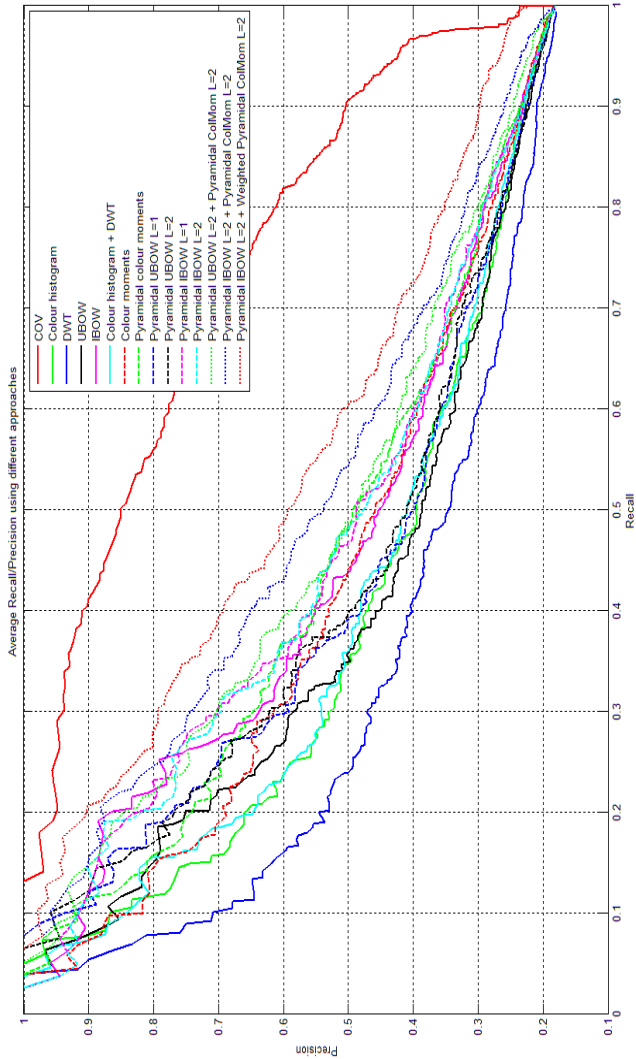

Figure 6: Recall-precision graph, for Vogel_6DS, of the 14 different approaches presented in Table 1 compared against the COV

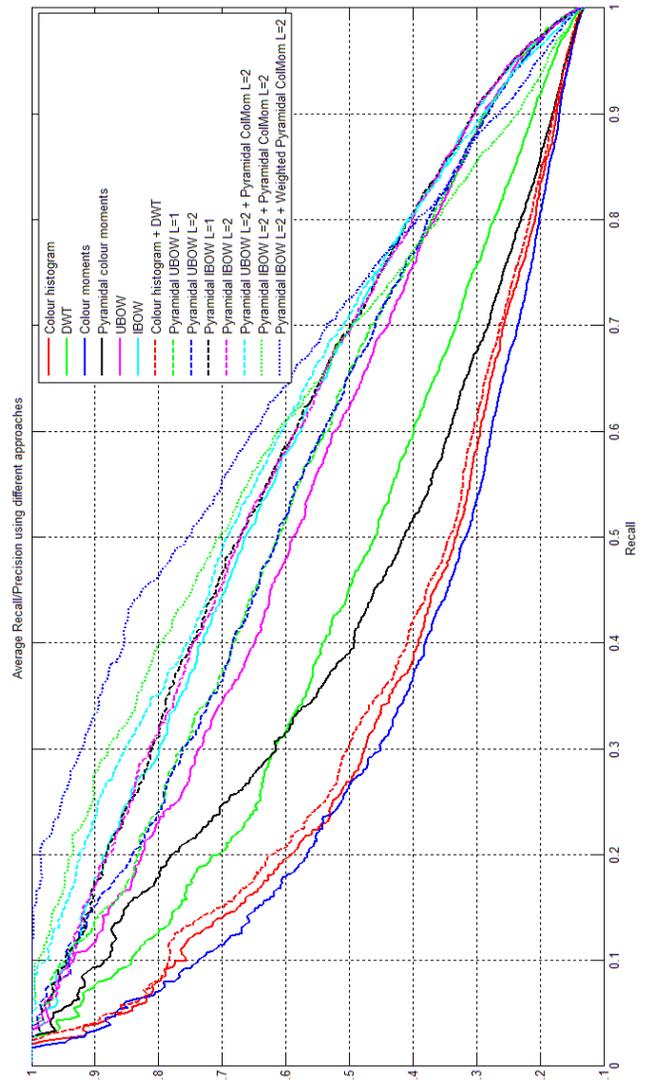

Figure 7: Recall-precision graph, for Oliva_8DS, of the 14 different approaches presented in Table 1.

### 5.5.3 Experimental results: Lazebnik_15DS

This section demonstrates the retrieval performance of using BOW-based approaches and other baseline methods evaluated on a larger dataset of 15 scene categories. The dataset contains gray images of Indoor and outdoor scenes. Some of the baseline approaches, presented in the previous two sections, are extracted from color images, i.e., from the three-color components of HSV. For gray images, these approaches are only available from one component, i.e., the gray component.

The experimental results show that the baseline methods failed to retrieve images for most of the scene categories. The color moments show good performance for the scene category Highway but failed for other scene categories. In contrast, the BOW-based approaches shown good retrieval performances compared with the baseline methods. The MAP for each scene category as well as the overall scene retrieval rate using different image representation approaches are listed in Table 6.

Table 6: The MAPs of each scene category, for Lazebnik_15DS, using the 14 different approaches presented in Table 1. The last column shows the retrieval accuracy of each of the corresponding approach.

| | \multicolumn{15}{c}{MAP per category} | | | | | | | | | | | | | | |
|---|---|---|---|---|---|---|---|---|---|---|---|---|---|---|---|
| | Suburban | Coast | Forest | Highway | Inside City | Mountain | Open country | Street | Tall building | Office | Bedroom | Industrial | Kitchen | Living room | Store | Accuracy |
| ColHist | 0.38 | 0.18 | 0.31 | 0.32 | 0.19 | 0.19 | 0.23 | 0.36 | 0.19 | 0.21 | 0.21 | 0.14 | 0.18 | 0.20 | 0.38 | 0.24 |
| PColMom_L0 | 0.20 | 0.25 | 0.23 | 0.42 | 0.20 | 0.20 | 0.24 | 0.28 | 0.23 | 0.18 | 0.18 | 0.21 | 0.18 | 0.18 | 0.29 | 0.23 |
| DWT | 0.39 | 0.51 | 0.63 | 0.52 | 0.30 | 0.30 | 0.40 | 0.40 | 0.39 | 0.40 | 0.25 | 0.19 | 0.23 | 0.32 | 0.43 | 0.38 |
| ColHist+DWT | 0.40 | 0.52 | 0.64 | 0.54 | 0.31 | 0.31 | 0.40 | 0.41 | 0.40 | 0.41 | 0.26 | 0.15 | 0.24 | 0.33 | 0.46 | 0.39 |
| PColMom_L2 | 0.37 | 0.28 | 0.25 | 0.50 | 0.25 | 0.23 | 0.32 | 0.48 | 0.26 | 0.19 | 0.18 | 0.22 | 0.19 | 0.19 | 0.36 | 0.28 |
| UBOW | 0.66 | 0.44 | 0.83 | 0.34 | 0.40 | 0.55 | 0.44 | 0.53 | 0.24 | 0.48 | 0.22 | 0.24 | 0.39 | 0.33 | 0.53 | 0.44 |
| IBOW | 0.63 | 0.57 | 0.88 | 0.43 | 0.49 | 0.58 | 0.41 | 0.54 | 0.36 | 0.53 | 0.22 | 0.27 | 0.38 | 0.29 | 0.51 | 0.47 |
| PUBOW_L1 | 0.69 | 0.45 | 0.88 | 0.34 | 0.40 | 0.53 | 0.49 | 0.53 | 0.24 | 0.46 | 0.22 | 0.26 | 0.37 | 0.32 | 0.55 | 0.45 |
| PUBOW_L2 | 0.69 | 0.45 | 0.90 | 0.34 | 0.41 | 0.52 | 0.49 | 0.53 | 0.24 | 0.45 | 0.23 | 0.26 | 0.37 | 0.32 | 0.57 | 0.45 |
| PUBOW_L2+PColMom_L2 | 0.70 | 0.43 | 0.88 | 0.46 | 0.42 | 0.47 | 0.54 | 0.57 | 0.24 | 0.43 | 0.23 | 0.25 | 0.37 | 0.31 | 0.57 | 0.46 |
| PIBOW_L1 | 0.66 | 0.58 | 0.89 | 0.42 | 0.50 | 0.56 | 0.43 | 0.58 | 0.36 | 0.54 | 0.22 | 0.27 | 0.36 | 0.29 | 0.49 | 0.48 |
| PIBOW_L2 | 0.66 | 0.57 | 0.89 | 0.43 | 0.50 | 0.55 | 0.43 | 0.59 | 0.37 | 0.54 | 0.22 | 0.26 | 0.34 | 0.29 | 0.49 | 0.48 |
| PIBOW_L2+PColMom_L2 | 0.54 | 0.52 | 0.82 | 0.58 | 0.45 | 0.38 | 0.50 | 0.60 | 0.33 | 0.42 | 0.22 | 0.25 | 0.31 | 0.26 | 0.50 | 0.44 |
| PIBOW_L2+WPColMom_L2 | 0.74 | 0.64 | 0.91 | 0.59 | 0.45 | 0.58 | 0.53 | 0.60 | 0.34 | 0.42 | 0.22 | 0.25 | 0.31 | 0.26 | 0.50 | 0.49 |

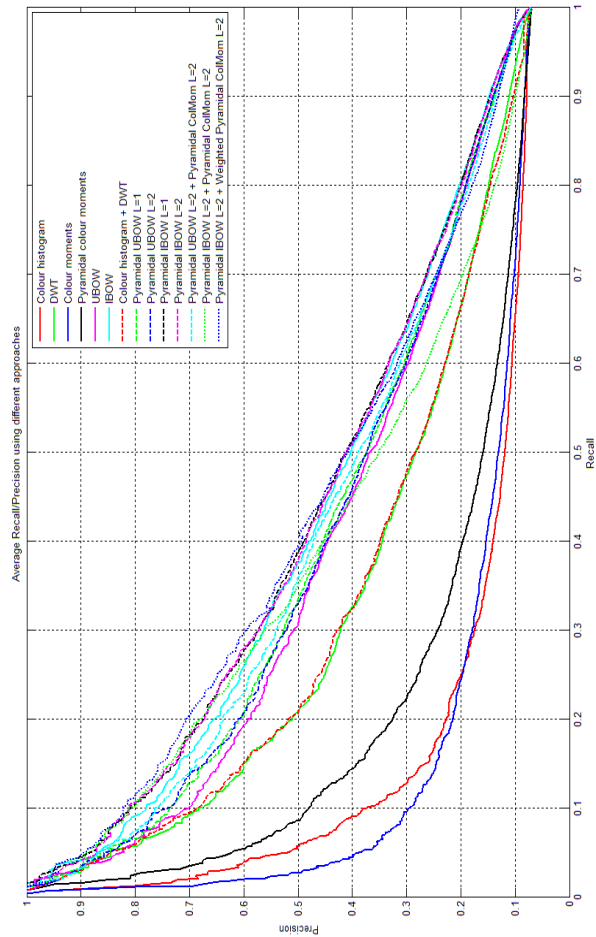

Figure 8: Recall-precision graph, for Lazebnik_15DS, of the 14 different approaches presented in Table 1.

The best retrieval rate is achieved using PIBOW_L2+WPColMom_L2. It indicates that the BOW-based approaches are appropriate for indoor/outdoor scene categories. The approach has reported (49%) retrieval rate with (+10%) increase in the retrieval rate over the best baseline methods. Also, the performance of the UBOW has achieved good results when the spatial pyramid layout is employed. The recall-precision graph of the different approaches is shown in Figure 8, where the recall-precision plots of each scene category are averaged for all different approaches.

## 6. CONCLUSION

In this paper, several experiments concerning the semantic retrieval of natural scenes have been carried out. The semantic representation of the natural scene images has been implemented using the annotated an un-annotated images. Firstly, the retrieval performance when employing the COV to summarize the amount of local semantic concepts depicted in an image have reported an encouraging results. The COV constructed from the labels of image regions represented by the BOW model have shown better performance compared with the baseline methods, such as color histogram, and also comparable with the COV benchmark. Secondly, the retrieval performance of using different configuration of the bag of visual word model have been studied and evaluated experimentally using three natural scene datasets. The experimental results obtained using the Vogel_6DS dataset have shown that the COV approaches achieved better retrieval accuracy compared to the BOW-based approaches and baseline. Also, the COV, as a global image representation, has lower dimensionality (9-D) than all other approaches. However, using the COV requires all image regions to be annotated manually. In the case of representing the semantic information of image content without using the COV approach, our previously proposed approaches presented in [33] for the image classification task have achieved better retrieval performance compared to other baseline methods on the datasets Oliva_8DS and Lazebnik_15DS. The retrieval accuracies of the two datasets using the different image representation approaches are reported.

**ACKNOWLEDGEMENTS**

The author is grateful to the Applied Science Private University, Amman, Jordan, for the full financial support granted to this research. The author would like to thank Dr. Julia Vogel for providing us access to the natural scene image dataset and for valuable discussion.